%% file: template.tex
\documentclass{article}

\usepackage{bm}
\usepackage{amsmath}

\usepackage{arxiv}

\usepackage[utf8]{inputenc} 
\usepackage[T1]{fontenc}    
\usepackage{hyperref}       
\usepackage{url}            
\usepackage{booktabs}       
\usepackage{amsfonts}       
\usepackage{nicefrac}       
\usepackage{microtype}      
\usepackage{lipsum}		
\usepackage{graphicx}
\usepackage{natbib}
\usepackage{doi}

\title{A 3D pocket-aware and affinity-guided diffusion model for lead optimization}

\date{} 					

\author{Anjie Qiao$^1$, Junjie Xie$^1$, Weifeng Huang$^3$, Hao Zhang$^3$, Jiahua Rao$^1$, \\ \textbf{Shuangjia Zheng$^{4\ast}$, Yuedong Yang$^{1\ast}$,  Zhen Wang$^{1\ast}$, Guo-Bo Li$^{2\ast}$, Jinping Lei$^{3\ast}$}\\
\\
\normalsize{$^{1}$School of Computer Science and Engineering, Sun Yat-sen University, 510006 Guangzhou, China}\\
\normalsize{$^{2}$West China School of Pharmacy, Sichuan University, 610041 Chengdu, China}\\
\normalsize{$^{3}$School of Pharmaceutical Science, Sun Yat-sen University, 510006 Guangzhou, China}\\
\normalsize{$^{4}$Global Institute of Future Technology, Shanghai Jiao Tong University, 200240 Shanghai, China}
\\ 
\normalsize{$^\ast$To whom correspondence should be addressed.}\\
\normalsize{E-mails: shuangjia.zheng@sjtu.edu.cn, yangyd25@mail.sysu.edu.cn,}\\ 
\normalsize{wangzh665@mail.sysu.edu.cn, liguobo@scu.edu.cn, leijp@mail.sysu.edu.cn.}
}

\hypersetup{
pdftitle={A template for the arxiv style},
pdfsubject={q-bio.NC, q-bio.QM},
pdfauthor={David S.~Hippocampus, Elias D.~Striatum},
pdfkeywords={First keyword, Second keyword, More},
}

\begin{document}
\maketitle

\begin{abstract}
	Molecular optimization, aimed at improving binding affinity or other molecular properties, is a crucial task in drug discovery that often relies on the expertise of medicinal chemists. Recently, deep learning-based 3D generative models showed promise in enhancing the efficiency of molecular optimization. However, these models often struggle to adequately consider binding affinities with protein targets during lead optimization. Herein, we propose a 3D pocket-aware and affinity-guided diffusion model, named Diffleop, to optimize molecules with enhanced binding affinity. The model explicitly incorporates the knowledge of protein-ligand binding affinity to guide the denoising sampling for molecule generation with high affinity. The comprehensive evaluations indicated that Diffleop outperforms baseline models across multiple metrics, especially in terms of binding affinity. 
\end{abstract}


\input{sections/introduction}
\input{sections/result}
\input{sections/methods}

\clearpage
\bibliographystyle{unsrtnat}
\bibliography{references}  






\end{document}

%% file: sections/introduction.tex
\section{ Introduction}
\label{sec.intro}
Molecular optimization, aims to improve the binding affinities of lead molecules, is a crucial and challenging step in the early stage of drug discovery. Traditionally, the optimization mostly depends on the medicinal chemists’ knowledge, experiences and intuitions~\cite{ref1,ref2,ref3}, which are inherently limited and not scalable or automated~\cite{ref4,ref5}. To aid and speed the traditional process, many fragment-based computational methods have been developed, including virtual fragment library construction and screening, fragment growing and linking, and scaffold hopping~\cite{ref6,ref7,ref8,ref9}. However, these methods were based on the computationally intuitive database search or physical simulations~\cite{ref10,ref11}, and the generated compounds have limited novelty and diversity~\cite{ref12}. 

Recent data-driven deep learning approaches have enabled alternative generative processes accelerate the structure-based molecular optimization~\cite{ref13,ref14,ref15,ref16,ref17}. Many target-aware deep generative models have been recently developed to capture the interactions between proteins and ligands for molecule generation with high affinity. Early approaches incorporate the protein pocket features as 1D sequence strings~\cite{ref18,ref19} or 2D pharmacophore graphs~\cite{ref20,ref21,ref22}, which are limited because the atom-level interactions between protein pockets and molecules are not explicitly trained and modeled~\cite{ref23,ref24}. This gave rise to the early attempts~\cite{ref25,ref26,ref27,ref28} of 3D pocket-aware generative models that represent the receptor-ligand complex as atomic density grids and employ 3D conventional neural network (CNN) to capture the geometric and spatial information of protein pocket. However, these 3D CNN models compressed the protein structure information, and it is hard to scale to large protein pockets~\cite{ref23,ref29}. To tackle this problem, the autoregressive generative models~\cite{ref23,ref29,ref30} based on 3D graph neural network (GNN) were proposed to learn the sequential conditional distribution~\cite{ref31} of different atoms in the 3D space around the binding pocket and generate molecules atom-by-atom inside the pocket. Nevertheless, this sequential generation method would lead to inaccurate and unreasonable 3D molecular structures because the position and element type of an atom are affected by all other atoms in the molecule~\cite{ref31,ref32}. Most recently, the 3D target-aware equivariant diffusion models that learn the joint distributions of all atoms within the 3D pocket have been proposed to generate all atoms of a molecule or fragment at once~\cite{ref11,ref31,ref32,ref33,ref34,ref35,ref36,ref37}. For example, DiffLinker~\cite{ref11} used an denoising diffusion model for fragment linking inside the 3D protein pocket. 

While the recently developed 3D target-aware diffusion-based generative models show effectiveness in generating molecules within specific protein pockets, they have limitations. First, these models neglect the explicit driving force of improving affinity for molecular optimization, and the incorporation of only 3D protein structural information is not adequate in guiding these models to generate optimized molecules with higher binding affinity~\cite{ref36}. Finally, most of these methods neglect chemical bond information in the training process, often causing unrealistic 3D molecular structures~\cite{ref42}. 

In this study, we have developed a new knowledge guided diffusion model Diffleop which incorporates affinity guidance and bond diffusion for molecular optimization with improved binding affinity. The binding affinity is conveyed to our model by fitting with an E(3)-equivariant neural network, and the bond diffusion is achieved through introduction of fake bond types and diffusion on fully connected molecular graphs. The experiments on generative performance revealed that Diffleop can optimize molecules with improved binding affinity and reasonable drug-like properties, and outperforms the baseline models on various evaluation metrics especially binding affinity. 

%% file: sections/result.tex
\section{Results}
\label{sec.results}
\subsection{Overview of the Diffleop model with affinity guidance}
\begin{figure}[t]
    \centering
    \includegraphics[width=0.85\linewidth]{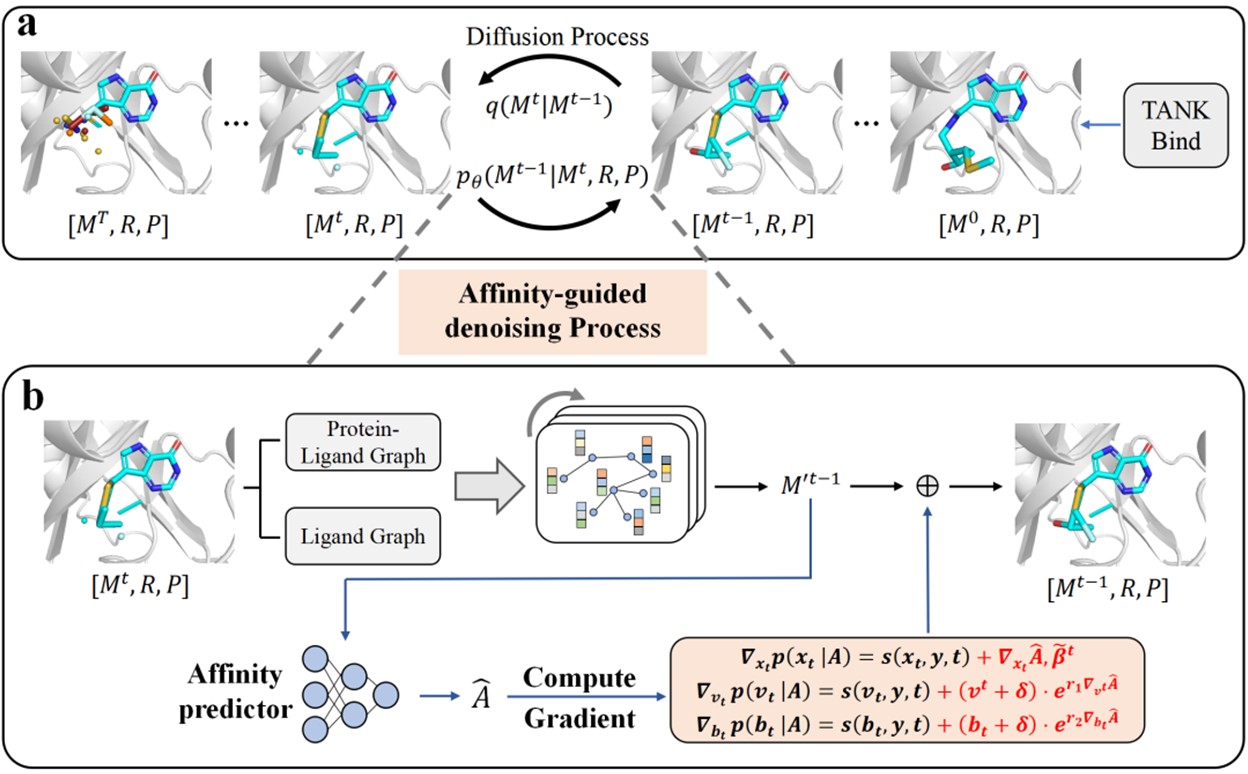}
    \caption{The architecture of Diffleop. (a) The framework of Diffleop. The diffusion process gradually adds predefined noise to the atom coordinates, atom types, and bond types of the ligand inside the protein pocket. The denoising process gradually reconstructs the molecules within the protein pocket from noisy ones. The protein structures ($P$) and retained groups of ligands ($R$, i.e. scaffolds or terminal fragments) are fixed during the whole diffusion and denoising process. The binding affinity between the intact molecule and protein is predicted by TANKBind. (b) $A$ detailed denoising procedure from time step $t$ to $t-1$ using equivariant graph neural network (EGNN) with affinity guidance, in which the trained affinity predictor calculates the binding affinity $\hat{A}$ and its gradients with respect to atom coordinates, atom types and bond types to guide the molecule generation toward high binding affinity.}
    \label{fig:framework}
\end{figure}
Diffleop is an affinity-guided diffusion-based generative model for 3D target-aware molecular optimization (Fig.~\ref{fig:framework}). This model consists of a forward diffusion process and a reverse denoising process, in which the protein pocket ($P$) and input fragments ($R$) are fixed and serve as the conditional information, and the covalent bond information is incorporated. During the diffusion process, the continuous Gaussian noises are gradually added at each time step to the atom coordinates of the fragment to be generated ($M$) through $T$ time steps, while the discrete noises characterized by categorical distribution are used for atom types and bond types of $M$. In the denoising procedure, the equivariant graph neural network (EGNN) was employed to invert the diffusion trajectory to generate atom coordinates, atom types and bond types simultaneously for reconstruction of the original molecule. The introducing of bond diffusion could eliminate the post-processing of bond inference and probably improve the quality of generated molecules.

The key contribution of Diffleop is the incorporation of binding affinity (Fig.~\ref{fig:framework}b) into the denoising process to guide the generation of molecules with improved binding affinity and favorable binding interactions. To achieve this, we firstly extracted molecule embeddings through EGNN and trained an affinity predictor to predict the binding affinities between the noised molecules and fixed proteins. Then, we utilized this trained affinity predictor to guide the molecule generation toward high affinity. At each denoising time step, the gradients of binding affinity $A$ with respect to atom coordinates, atom types and bond types are calculated to provide the direction of improving binding affinity and guide the determination of atom positions, atom types and bond types. Thus, the desired molecules with enhanced binding affinities could be generated by implementing the guidance on atom coordinates, atom and bond types across the whole denoising process.

\subsection{Diffleop is effective for molecular optimization with enhanced binding affinity and outperforms baseline models}
\begin{figure}[h]
    \centering
    \includegraphics[width=0.9\linewidth]{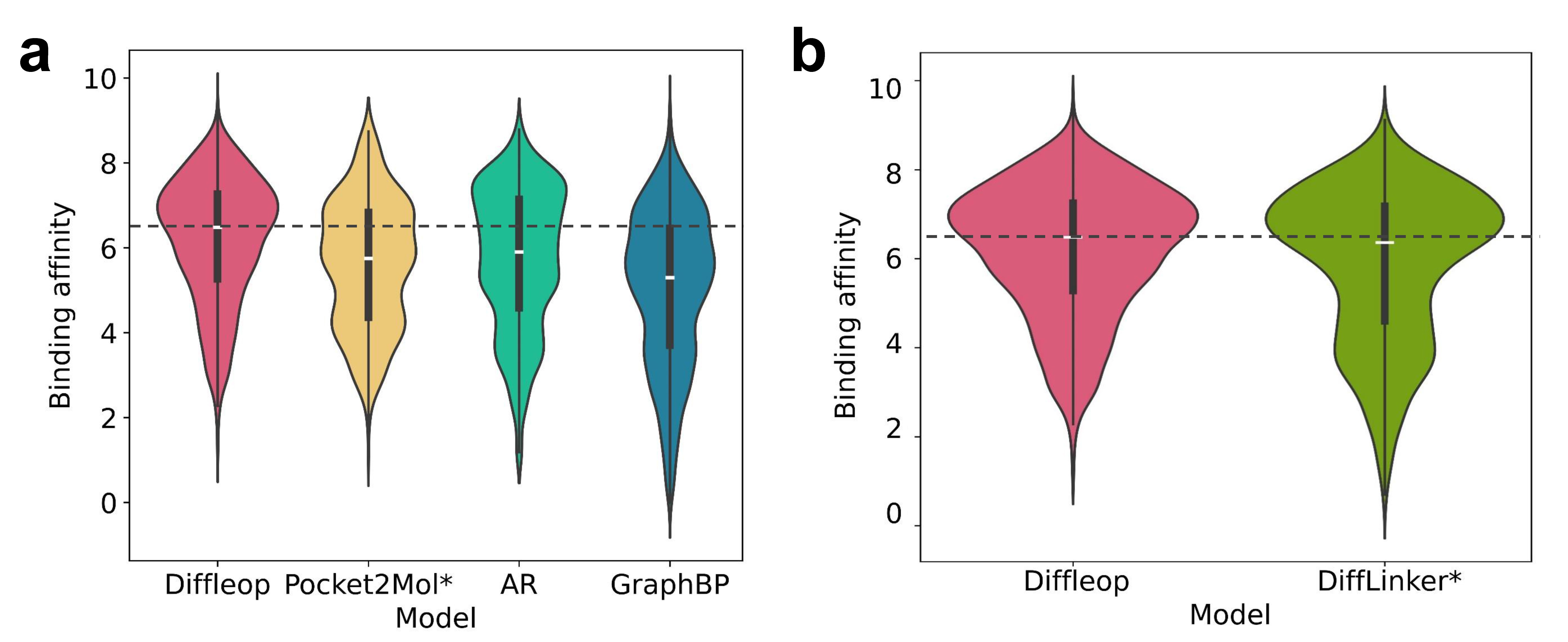}
    \caption{Diffleop achieves state-of-art performance on molecular optimization through scaffold decoration and linker design tasks. ‘*’ means the models are retrained on our training set.}
    \label{fig:result}
\end{figure}

We evaluated and compared the molecular optimization performance of Diffleop with five 3D pocket-aware generative models including GraphBP~\cite{ref29}, AR~\cite{ref30}, Pocket2Mol~\cite{ref23}, and DiffLinker~\cite{ref11} through a range of metrics, such as binding affinity (Affinity), drug-likeness (QED), synthetic accessibility (SA), LogP (octanol-water partition coefficient), Lipinski's rule-of-five (Table~\ref{table1}, Fig.~\ref{fig:result}). In addition, we performed ablation experiments to elucidate the effectiveness and importance of affinity guidance and bond diffusion on our Diffleop model (Table~\ref{Ablation}). For each pair of scaffold/fragments and pockets in the test set with 100 protein targets, we generated 100 molecules for assessment. The binding affinities of the generated 3D molecules were calculated by TANKBind~\cite{ref39}, and the top 5 molecules with highest binding affinities were selected for binding-related metrics. The common drug-like properties (QED, SA, LogP, Lipinski's rule-of-five) were calculated by using RDKit~\cite{ref44}.

Diffleop significantly outperforms all state-of-the-art baseline models on all binding-related metrics including Affinity and High Affinity (Table~\ref{table1}, Table~\ref{table2}, Fig.~\ref{fig:result}). Herein, Affinity is the most important binding-related metric as it directly evaluates the binding affinity of generated 3D molecules without optimizing the conformation. Diffleop shows significant higher average of Affinity than all baseline models, and outperforms reference that directly determined from the ligand-protein pairs in test set. 

\begin{table}[h]
\centering
\begin{tabular}{ccccccc}
\bottomrule
Method                           & Affinity & High Affinity, \% & QED   & SA    & LogP  & Lipinski \\
\bottomrule
\multicolumn{1}{c|}{Reference}   & 6.69     & -                 & 0.642 & 0.770 & 3.085 & 4.92     \\
\multicolumn{1}{c|}{GraphBP}     & 5.47     & 17.80             & 0.605 & 0.316 & 0.567 & 4.82     \\
\multicolumn{1}{c|}{AR}          & 5.93     & 18.60             & 0.660 & 0.710 & 1.855 & 4.91     \\
\multicolumn{1}{c|}{Pocket2Mol*} & 5.78     & 21.80             & 0.704 & 0.773 & 1.904 & 4.96     \\
\multicolumn{1}{c|}{Diffleop}    & 6.71     & 52.60             & 0.654 & 0.706 & 2.772 & 4.93     \\ 
\bottomrule
\end{tabular}
\caption{ Performance comparison between Diffleop and baseline models on molecular optimization through scaffold decoration.}
\label{table1}
\end{table}

The evaluated common properties (Table ~\ref{table1}, Table~\ref{table2}) indicated that Diffleop generates molecules with reasonable drug-like properties and shows comparable performance with the state-of-the-art baseline models AR, GraphBP, Pocket2Mol, and DiffLinker. Specifically, the average LogP value of molecules generated by Diffleop falls within the compliance range ($-0.4 \sim 5.6$) and close to those of reference ligands, while molecules generated by baseline models exhibit lower LogP values with higher hydrophilicity. For synthetic accessibility metric SA, molecules generated by Diffleop shows comparable results to the reference ligands and those generated by AR and Pocket2Mol, while molecules generated by GraphBP and DiffLinker exhibit significantly lower SA values, indicating that molecules generated by Diffleop are more likely easy to synthesize than those generated by GraphBP and DiffLinker. For drug-likeness properties QED and Lipinski's rule, molecules generated by Diffleop exhibit comparable results with baseline models and slightly higher average value than reference ligands, indicating Diffleop can generate molecules with better drug-like properties than reference ligands.

Therefore, Diffleop is capable of generating more drug-like molecules with higher binding affinity and ligand efficiency than the baseline models. This reveals the effectiveness of affinity guidance and bond diffusion employed in Diffleop, while the baseline models solely depend on learning the joint distribution between molecular scaffolds/fragments and the protein pocket. The ablation experiments (Table~\ref{Ablation}) showed that Diffleop without affinity guidance achieves significantly lower average binding affinity and lower percentage of High Affinity than the original model, further indicating the effect of affinity guidance for the determination of atom positions, atom types and bond types in the sampling process of Diffleop for molecular generation with enhanced affinity. In addition, the molecules generated by Diffleop without bond diffusion exhibit decreased drug-like properties, i.e. lower QED, SA, and Lipinski's values than the original model. Hence, the bond diffusion in Diffleop is capable of learning not only the distribution of bond types but also the implicit chemical properties of the structures. Consequently, Diffleop could generate more drug-like molecules with more rational structures.

\begin{table}[h]
\centering
\begin{tabular}{ccccccc}
\bottomrule
Method                           & Affinity & High Affinity, \% & QED   & SA    & LogP  & Lipinski \\ \bottomrule
\multicolumn{1}{c|}{Reference}   & 6.73     & -                 & 0.601 & 0.770 & 2.867 & 4.84     \\
\multicolumn{1}{c|}{DiffLinker}  & 6.09     & 26.87             & 0.625 & 0.321 & 0.893 & 4.91     \\
\multicolumn{1}{c|}{DiffLinker*} & 6.14     & 28.89             & 0.635 & 0.321 & 0.948 & 4.93     \\
\multicolumn{1}{c|}{Diffleop}    & 6.74     & 54.60             & 0.604 & 0.716 & 2.865 & 4.85     \\ 
\bottomrule
\end{tabular}
\caption{ Performance comparison between Diffleop and DiffLinker~\cite{ref11} on linker design task. * The method was retrained on our training set.}
\label{table2}
\end{table}

\begin{table}[h]
\centering
\begin{tabular}{c|cccccc}
\bottomrule
Method                    & Binding Affinity & High Affinity (\%) & QED   & SA    & LogP  & Lipinski \\ \bottomrule
                          & \multicolumn{6}{c}{Scaffold Decoration}                                  \\ \bottomrule
Without all               & 6.49             & 36.80              & 0.677 & 0.694 & 2.612 & 4.93     \\
Without bond diffusion    & 6.71             & 55.60              & 0.630 & 0.664 & 2.817 & 4.89     \\
Without affinity guidance & 6.49             & 40.40              & 0.697 & 0.735 & 2.693 & 4.95     \\
Diffleop                  & 6.71             & 52.60              & 0.654 & 0.706 & 2.772 & 4.93     \\ \bottomrule
                          & \multicolumn{6}{c}{Linker Design}                                        \\ \bottomrule
Without all               & 6.59             & 42.20              & 0.615 & 0.686 & 2.643 & 4.88     \\
Without bond diffusion    & 6.83             & 58.40              & 0.554 & 0.638 & 2.682 & 4.76     \\
Without affinity guidance & 6.60             & 44.00              & 0.617 & 0.728 & 2.844 & 4.88     \\
Diffleop                  & 6.74             & 54.60              & 0.604 & 0.716 & 2.865 & 4.85     \\ \bottomrule
\end{tabular}
\caption{ Ablation study of bond diffusion and affinity guidance on scaffold decoration and linker design tasks.}
\label{Ablation}
\end{table}

%% file: sections/methods.tex
\section{Materials and Methods}
\label{sec.methods}
\subsection{Preliminary}
The protein pocket can be depicted as a set of $N^P$ atoms $\bm{P} = \left\{ \left( \bm{x}_i^P, \bm{v}_i^P \right) \right\}_{i\in \{1,\ldots,N^P\}}$. 
Similarly, the ligand molecule can be described as a set of $N^L$ atoms $\bm{L} = \left\{ \left( \bm{x}_i^L, \bm{v}_i^L, \bm{b}_{ij}^L \right) \right\}_{i,j\in \{1,\ldots,N^L\}}$. 
In scaffold decoration and linker design, the scaffold and two terminal fragments are respectively given as conditions along with the protein pocket. 
Therefore, we collectively denote the scaffolds and terminal fragments, which act as context along with the protein pocket, as the "retain" $\bm{R} = \left\{ \left( \bm{x}_i^R, \bm{v}_i^R, \bm{b}_{ij}^R \right) \right\}_{i,j\in \{1,\ldots,N^L\}}$, and the fragment to be generated as the "mask" $\bm{M} = \left\{ \left( \bm{x}_i^M, \bm{v}_i^M, \bm{b}_{ij}^M \right) \right\}_{i,j\in \{1,\ldots,N^M\}}$. 
Here, $\bm{x}_i \in \mathbb{R}^3$, $\bm{v}_i \in \mathbb{R}^d$ and $\bm{b}_{ij} \in \mathbb{R}^{d'}$ represent the 3D geometric coordinate, the element type of the $i$-th atom, and the bond type between the $i$-th and $j$-th atom, respectively.

\subsection{Diffusion Process}
We perform diffusion on atomic coordinates, atomic types, and bond types separately. For atomic coordinates, we introduce noise iteratively to the atoms within the mask $\bm{M}^0$ by drawing from a Gaussian distribution $\mathcal{N}$ at each time step, and keep the coordinates of ``retain'' and protein components unchanged. For atom types and bond types, we employ categorical distribution $\mathcal{C}$ due to their discrete characteristics~\cite{ref58}. Given the state at time step $t-1$, the conditional distribution for time step $t$ can be expressed as follows:

\[
q(x^t|x^{t-1}) = \mathcal{N}(x^t| \sqrt{ \bar{\alpha}^t } x^{t-1}, \bar{\sigma}^t \bm{I}), \tag{1}
\]
\[
q(v^t|v^{t-1}) = \mathcal{C}(v^t| \sqrt{\bar{\alpha}^t} v^{t-1} + \bar{\sigma}^t/K_v), \tag{2}
\]
\[
q(b^t|b^{t-1}) = \mathcal{C}(b^t| \sqrt{\bar{\alpha}^t} b^{t-1} + \bar{\sigma}^t/K_b), \tag{3}
\]
where $\sqrt{{\bar{\alpha}}^t} \in \mathbb{R}^+$ represents the amount of original signal, $\sqrt{{\bar{\sigma}}^t} \in \mathbb{R}^+$ denotes the part of noise to add, $\bm{I} \in \mathbb{R}^{3\times3}$ is the identity matrix, $K_v$ and $K_b$ stand for the number of atom types and bond types, respectively.

Leveraging the inherent Markov property in the diffusion process, the entire noising process can be formulated as:

\[
q(M^1, M^2, \ldots, M^T|M^0) = \prod_{t=0}^T q(M^t|M^{t-1}). \tag{4}
\]

The distribution $q$ is independent at each time step, we can derive the distribution of $\bm{M}^t$ based on $\bm{M}^0$ as follows:

\[
q(x^t|M^0) = \mathcal{N}(x^t| \sqrt{\alpha^t} x^0, \sigma^t \bm{I}), \tag{5}
\]
\[
q(v^t|M^0) = \mathcal{C}(v^t| \sqrt{\alpha^t} v^0 + \sigma^t/K_v), \tag{6}
\]
\[
q(b^t|M^0) = \mathcal{C}(b^t|\sqrt{\alpha^t} b^0 + \sigma^t/ K_b), \tag{7}
\]
where ${\bar{\alpha}}^t = \alpha^t/\alpha^{t-1}$ and ${\bar{\sigma}}^t = \sigma^t - {\bar{\alpha}}^t {\bar{\sigma}}^{t-1}$.
Following Kingma et al.~\cite{ref59}, we calculate the signal-to-noise ratio as $SNR(t) = \alpha^t/\sigma^t$. Typically, $\sqrt{\alpha^t}$ evolves according to a learned or pre-defined schedule from $\sqrt{\alpha^0} \approx 1$ towards $\sqrt{\alpha^T} \approx 0$~\cite{ref59}. Alternatively, a variance-preserving diffusion approach can be adopted where $\sqrt{\alpha^t} = \sqrt{1-\sigma^t}$~\cite{ref60}. This closed-form formula demonstrates that we can achieve the noisy data distribution at any time step without the need for iterative computations.

Following molecule size prediction strategies in DiffDec~\cite{ref37}, we integrate the fake atom mechanism to incorporate the mask size prediction into the diffusion model. In addition, we also add the fake bond to bond types to introduce bond diffusion into our model, aiming to eliminate the post-processing of bond inference after the generation of coordinates and atomic types in DiffDec and DiffLinker.

\subsection{Denoising Process}
In the denoising procedure, we invert the Markov chain to recover the original sample from prior distributions within the fixed context of ``retain'' and protein pocket. The transition distribution is formulated as:

\[
p_\theta(M^{t-1}|M^t,R,P) = q(M^{t-1}|\hat{M}^0, M^t, R, P), \tag{8}
\]
where $p_\theta$ is a neural network to parameterize the transition.

We instruct the neural network to reconstruct $\bm{M}^{t-1}$ from $\bm{M}^t$ by optimizing the predicted distributions $p_\theta\left(\bm{M}^{t-1}\middle|\bm{M}^t,R,P\right)$ to approach the real posterior $q\left(\bm{M}^{t-1}\middle|\bm{M}^t,\bm{M}^0\right)$. Both $\bm{M}^{t-1}$ and $\bm{M}^t$ can be obtained in the forward process. The loss functions are defined as follows:

\[
\mathcal{L}^{t-1} = \mathcal{L}^{t-1}_{\text{pos}} + \lambda_1 \mathcal{L}^{t-1}_{\text{atom}} + \lambda_2 \mathcal{L}^{t-1}_{\text{bond}}, \tag{9}
\]
\[
\mathcal{L}^{t-1}_{\text{pos}} = \frac{1}{N} \sum_i \left\| x_i^{t-1} - \mu_\theta(M^t, t)_i \right\|_2^2, \tag{10}
\]
\[
\mathcal{L}^{t-1}_{\text{atom}} = \frac{1}{N} \sum_i \mathrm{D_{KL}}(q(v_i^{t-1}|M^t,M^0) \,\|\, p_\theta(v_i^{t-1}|M^t)), \tag{11}
\]
\[
\mathcal{L}^{t-1}_{\text{bond}} = \frac{1}{N^2}\sum_{ij}  \mathrm{D_{KL}}(q(b_{ij}^{t-1}|M^t,M^0) \,\|\, p_\theta(b_{ij}^{t-1}|M^t)), \tag{12}
\]
where $\lambda_1$ and $\lambda_2$ are predefined constants.

In the training phase, a time step $t$ is randomly sampled, and the neural network is optimized by minimizing the loss function. Then, to generate new molecules, we initially sample $\bm{M}^t$ from the prior distribution and subsequently sample $\bm{M}^{t-1}$ from $p_\theta\left(\bm{M}^{t-1}\middle|\bm{M}^t\right)$ for $t=T,T-1,\ldots,1$ to gradually denoise the output.

For scaffold hopping, we firstly add noise for $t$ steps through the forward diffusion process to the specific fragment for optimization. Subsequently, we denoise from the $T-t$th step in the molecular generation process.

\subsection{Equivariant Graph Neural Networks}
We employ the equivariant graph neural network (EGNN)~\cite{ref58,ref61,ref62} with nodes and edges updating to denoise the 3D molecular graph. Additionally, we integrate construction of protein-ligand graph and ligand graph within the network and conduct message passing to enhance the atomic and bond representation of a molecule within the protein pocket. Specifically, the protein-ligand graph is a k-nearest neighbors (knn) graph $G_K$ upon protein atoms and ligand atoms to capture the interactions between the protein and ligand:

\[
\Delta h_{K,i} \leftarrow \sum_{j\in \mathcal{N}_K(i)} \phi_{m_K}(h_i,h_j, 
|| x_i-x_j||,E_{ij},t), \tag{13}
\]
where $\bm{h}$ is the atom's hidden state, $\mathcal{N}_K(i)$ denotes the neighbors of atom $i$ in $G_K$, and $E_{ij}$ specifies whether the edge between atom $i$ and $j$ corresponds to a protein-protein, ligand-ligand or protein-ligand edge.

On the other hand, the ligand graph is a fully connected graph $G_L$ that incorporates the bond types among ligand atoms to depict the internal interaction within the ligand:

\[
m_{ij} \leftarrow \phi_d(||x_i-x_j||,e_{ij}), \tag{14}
\]
\[
\Delta h_{L,i} \leftarrow \sum_{j\in \mathcal{N}_L(i)} \phi_{m_L}(h_i,h_j,m_{ji},t), \tag{15}
\]
where $\bm{e}$ is the bond's hidden state. The atom's hidden state is updated based on the aggregated messages:

\[
\Delta h_i \leftarrow h_i + \phi_h(\Delta h_{K,i} + \Delta h_L,i). \tag{16}
\]
The bond's hidden state is updated following a directional message passing~\cite{ref63,ref64} scheme:

\[
e_{ji} \leftarrow \sum_{k\in \mathcal{N}_L(j)\setminus i} \phi_e(h_i,h_j,h_k,m_{kj},m_{ji},t). \tag{17}
\]

Finally, the atom positions of the ligand are updated:

\[
\Delta x_{K,i} \leftarrow \sum_{j\in \mathcal{N}_K(i)} (x_j-x_i) \phi_{x_K}(h_i,h_j,||x_i-x_j||,t), \tag{18}
\]
\[
\Delta x_{L,i} \leftarrow \sum_{j\in \mathcal{N}_L(i)} (x_j-x_i) \phi_{x_L}(h_i,h_j,||x_i-x_j||,m_{ji},t), \tag{19}
\]
\[
x_i \leftarrow x_i + (\Delta x_{K,i} + \Delta x_{L,i}) \cdot \mathbf{1}_{\text{mask}}, \tag{20}
\]
where $\mathbf{1}_{\text{mask}}$ is the indicator of masked atoms in the ligand.

The final hidden states $\bm{h}^M$ and $\bm{e}^L$ are fed into two MLPs to obtain the predicted atom type $\hat{\bm{v}}_i = \mathrm{softmax}(\mathrm{MLP}(\bm{h}_i^M))$ and bond type $\hat{\bm{b}}_{ij} = \mathrm{softmax}(\mathrm{MLP}(\bm{e}_{ij}^M+\bm{e}_{ji}^M))$.

\subsection{Affinity Guidance in Denoising Process}
We directly incorporate the binding affinity of the protein-ligand complex in the denoising process. Firstly, we utilize the molecule embeddings extracted by EGNN and train an affinity predictor to predict the binding affinity $\hat{A}$ between the noised ligand $\bm{L}^t$ and the protein target $\bm{P}$ through an MLP:

\[
A = \frac{1}{N_L} \sum_{i=0}^{N_L-1} \mathrm{sigmoid}(\mathrm{MLP}(L^t,P,t)). \tag{21}
\]

Then in the denoising process, we utilize the trained affinity predictor to guide the molecule generation. At time step $t$, the neural network takes the protein-ligand complex as inputs, reconstructs the molecule $\bm{L}^t$, predicts the binding affinity $\hat{A}$, and calculates the gradient of atom coordinates, atom types and bond types to guide the determination of atomic positions, atomic and bond types~\cite{ref65}.

For atom coordinates $\bm{x}^t$, the gradient of $\hat{A}$ concerning $\bm{x}^t$ offers guidance on where to place the atoms to improve the binding affinity, and $\bm{x}^{t-1}$ can be effectively sampled from a conditional Gaussian transition process as follows:

\[
x^{t-1} \sim \mathcal{N}\left( \widetilde{\mu}^t(x^t,\hat{x}^0) + s\widetilde{\beta}^t \mathbf{I} \nabla_{x^t} \hat{A}, \widetilde{\beta}^t \mathbf{I} \right), \tag{22}
\]
where $s$ is an atom coordinate gradient scale, $\widetilde{\mu}^t(x^t,\hat{x}^0) = \frac{\sqrt{\bar{\alpha}^t}\sigma^{t-1}}{\sigma^t}x^t + \frac{\sqrt{\alpha^{t-1}}\bar{\sigma}^t}{\sigma^t}\hat{x}^0$ and $\widetilde{\beta}^t = \frac{\bar{\sigma}^t\sigma^{t-1}}{\sigma^t}$.

For atom types $\bm{v}$, the gradient of $\hat{A}$ relative to $\bm{v}^t$ is calculated by applying the one-hot encoding of atom types. A positive gradient increases the values within the vector, while a negative gradient leads the values towards zero. To allow the gradients to impact the positions of zero, we introduce a slight positive value $\bm{\delta}$ across the entire one-hot vector. The guidance for $\bm{v}^t$ is depicted as follows:

\[
v^t \leftarrow (v^t + \delta) \cdot e^{r_1} \nabla_{v^t} \hat{A}, \tag{23}
\]
where $r_1$ is an atom type gradient scale.

For bond types $\bm{b}$, we use the same mechanism as that in atom types, and the guidance for $\bm{b}^t$ is given as follows:

\[
b^t \leftarrow (b^t + \delta) \cdot e^{r_2} \nabla_{b^t} \hat{A}, \tag{24}
\]
where $r_2$ is a bond type gradient scale.

\subsection{Baselines}
We compare the performance of our Diffleop method with the following 3D pocket-aware generative baseline models:

\begin{itemize}
    \item \textbf{GraphBP}~\cite{ref29}. A 3D auto-regressive model for generation of 3D molecules that bind to specific proteins by sequentially placing atoms through predicting the internal coordinates for structure-based de novo drug design.
    \item \textbf{AR}~\cite{ref30}. A 3D auto-regressive generative model for structure-based de novo drug design, in which the atom types and coordinates are sequentially sampled through MCMC strategy.
    \item \textbf{Pocket2Mol}~\cite{ref23}. An E(3)-equivariant auto-regressive model for structure-based de novo drug design that creates 3D molecular structures by sequentially placing atoms through predicting atom types and relative 3D coordinates.
    \item \textbf{DiffLinker}~\cite{ref11}. An E(3)-equivariant 3D-conditional diffusion model for molecular linker design, which also incorporates 3D protein pocket constraints without domain knowledge guidance and only employs diffusion on atom types and coordinates. This model servers an effective tool to generate valid linkers conditioned on target protein pockets.
\end{itemize}

\subsection{Evaluation Metrics}
We access the quality of generated molecules with a range of metrics:

\begin{itemize}
    \item \textbf{Affinity}: the binding affinity between the generated 3D molecules and the target pocket predicted by TANKBind~\cite{ref39}.
    \item \textbf{High affinity}: the percentage of the top 5 generated molecules whose binding affinity is higher than or equal to that of the reference compound.
    \item \textbf{QED}~\cite{ref66}: the quantitative estimation of drug-likeness, assessing molecules' potential to become a viable drug candidate.
    \item \textbf{SA}~\cite{ref67}: the synthetic accessibility score that evaluates the difficulty of synthesizing a compound. It is normalized between 0 and 1, where a higher value means easier synthesizability.
    \item \textbf{LogP}~\cite{ref68}: the octanol-water partition coefficient, whose values generally fall within the range of -0.4 to 5.6 for good drug candidates.
    \item \textbf{Lipinski}~\cite{ref69}: how many rules the generated molecule follows the Lipinski’s rule-of-five.
\end{itemize}